%% file: partgen.tex
\pdfoutput=1

\documentclass[10pt,twocolumn,letterpaper]{article}

\usepackage{cvpr}
\usepackage{times}
\usepackage{epsfig}
\usepackage{graphicx}
\usepackage{amsmath}
\usepackage{amssymb}
\usepackage{makecell}
\usepackage{multirow}
\input{ourcommands}

\usepackage[pagebackref=true,breaklinks=true,letterpaper=true,colorlinks,bookmarks=false]{hyperref}

\cvprfinalcopy 

\def\cvprPaperID{5994} 

\ifcvprfinal\fi
\begin{document}

\title{PQ-NET: A Generative Part Seq2Seq Network for 3D Shapes}

\author{
	Rundi Wu$^{1,4}$~~~~~~
	Yixin Zhuang$^{1}$~~~~~~
	Kai Xu$^{2}$~~~~~~
	Hao Zhang$^{3,4}$~~~~~~
	Baoquan Chen$^{1,4}$~~~~~~
	\\
	$^1$Center on Frontiers of Computing Studies, Peking University\\
	$^2$National University of Defense Technology\\
	$^3$Simon Fraser University \quad\quad 
	$^4$AICFVE, Beijing Film Academy
}

\maketitle

``The characterization of object perception provided by {\em recognition-by-components\/} (RBC) bears a close
resemblance to some current views as to how {\em speech\/} is perceived.''
\vspace{-15pt}
\begin{flushright}
--- Irving Biederman~\cite{RBC1987}
\end{flushright}

\input{abstract}

\input{intro}

\input{related}

\input{method}

\input{results}

\input{application}

\input{conclusion}

\vspace{-5pt}
\section*{Acknowledgement}
We thank the anonymous reviewers for their valuable comments. This work was supported in part by National Key R\&D Program of China (2019YFF0302902), NSFC (61902007), NSERC Canada (611370), and an Adobe gift. Kai Xu is supported by National Key R\&D Program of China (2018AAA0102200).

{\small
\bibliographystyle{ieee}
\bibliography{partgen}
}

\newpage
\input{supp}

\end{document}

%% file: ourcommands.tex
\usepackage{color}
\definecolor{green}{rgb}{0, 0.5, 0}
\definecolor{orange}{rgb}{0.8, 0.6, 0.2}
\definecolor{red}{rgb}{1.0, 0.0, 0.0}
\definecolor{teal}{rgb}{0.0, 0.4, 0.4}
\definecolor{purple}{rgb}{0.65,0,0.65}
\definecolor{saffron}{rgb}{0.95,0.75,0.2}
\definecolor{turquoise}{rgb}{0.0,0.5,0.5}
\definecolor{black}{rgb}{0.0, 0.0, 0.0}

\usepackage[normalem]{ulem}

\usepackage{amsopn}
\usepackage{mathtools}
\usepackage{amsmath, amsthm, amsfonts, amssymb}
\usepackage{graphicx} 
\usepackage{textcomp}
\usepackage{xfrac}

\usepackage{overpic}
\usepackage{graphicx}
\usepackage{float}
\usepackage{multirow}


%% file: abstract.tex

\begin{abstract}
We introduce PQ-NET, a deep neural network which represents and generates 3D shapes via {\em sequential part assembly\/}.
The input to our network is a 3D shape segmented into parts, where each part is first encoded into a feature representation using 
a part autoencoder. The core component of PQ-NET is a {\em sequence-to-sequence\/} or {\em Seq2Seq\/} autoencoder which 
encodes a sequence of part features into a latent vector of fixed size, and the decoder reconstructs the 3D shape, one part 
at a time, resulting in a sequential assembly.
%
%
The latent space formed by the Seq2Seq encoder encodes both part structure and fine part geometry. 
The decoder can be adapted to perform several generative tasks including shape autoencoding, interpolation, novel shape 
generation, and single-view 3D reconstruction, where the generated shapes are all composed of meaningful parts.  
\end{abstract}

%% file: intro.tex

\section{Introduction}
\label{sec:intro}

Learning generative models of 3D shapes is a key problem in both computer vision and computer graphics. While graphics is mainly
concerned with 3D shape modeling, in inverse graphics~\cite{hinton2012}, a major line of work
in computer vision, one aims to infer, often from a single image, a disentangled representation with respect to 3D shape 
and scene structures~\cite{kulkarni2015}. Lately, there has been a steady stream of works on developing deep neural networks for 3D shape 
generation using different shape representations, e.g., voxel grids~\cite{wu2016learning}, point clouds~\cite{fan2016point,Achlioptas2018}, 
meshes~\cite{groueix2018papier,wang2018pixel2mesh}, and most recently, implicit functions~\cite{OccNet,park2019deepsdf,chen2018implicit_decoder,DISN}.
However, most of these works produce {\em unstructured\/} 3D shapes, despite the fact that object perception is generally believed to be 
a process of {\em structural understanding\/}, i.e., to infer shape parts, their compositions, and inter-part relations~\cite{hoffman1984,RBC1987}.

\begin{figure}[t] \centering
    \includegraphics[width=0.99\linewidth, height=0.68\linewidth]{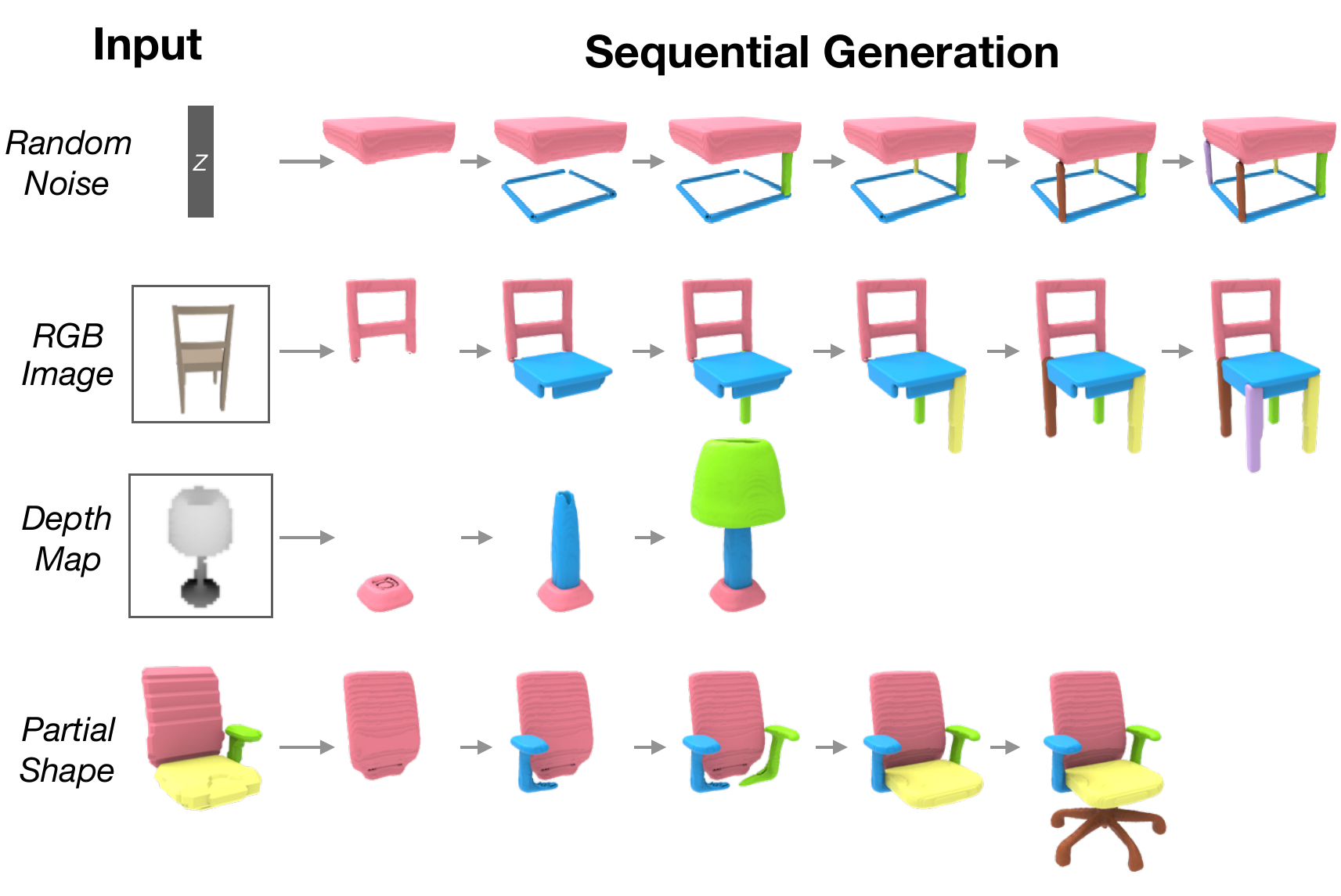}
    \caption{Our network, PQ-NET, learns 3D shape representations as a {\em sequential part assembly\/}. It can be adapted to generative tasks such as random 3D shape generation,
    single-view 3D reconstruction (from RGB or depth images),  and shape completion.}
    \label{fig:teaser}
\end{figure}

\begin{figure*}
\begin{center}
{\includegraphics[width=1.0\linewidth, height=0.42\linewidth]{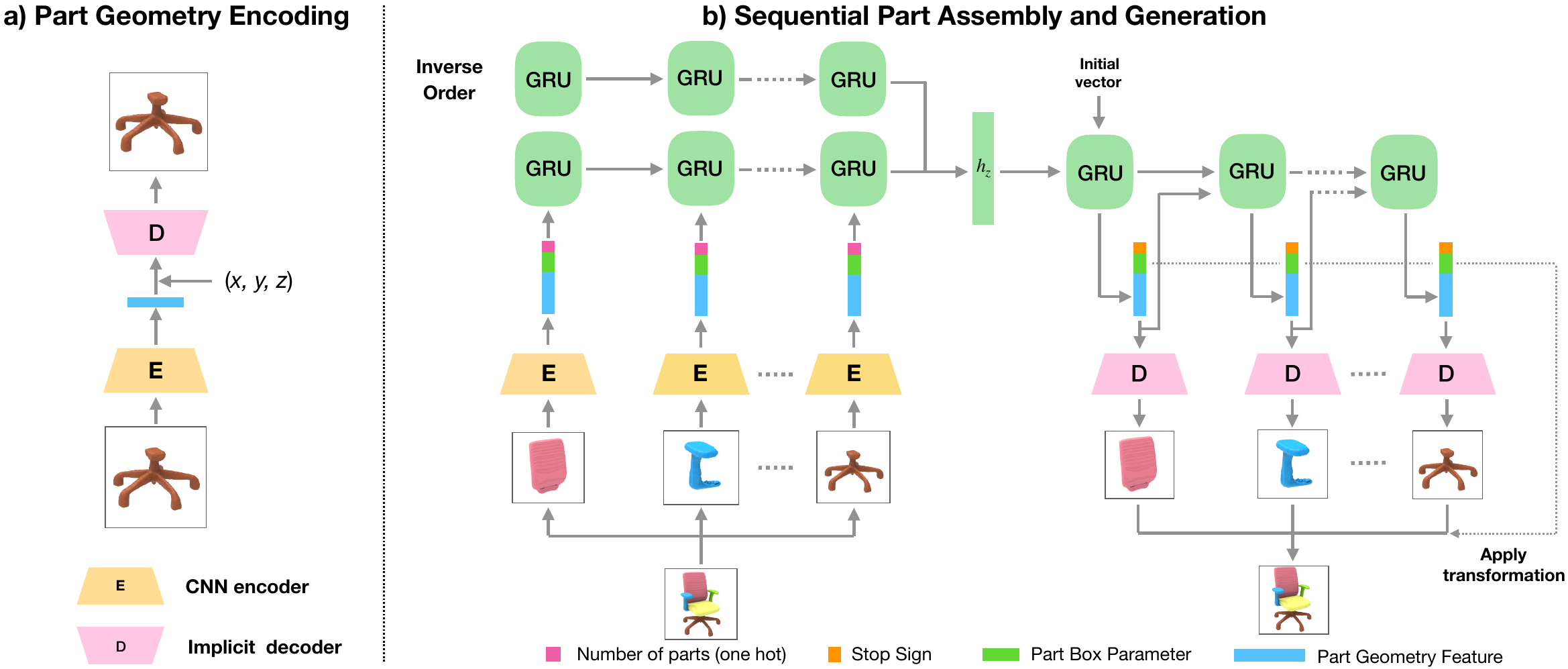} }
\end{center}
   \caption{The architecture of PQ-NET: our part Seq2Seq generative network for 3D shapes.}
\label{fig:architecture}
\end{figure*}

In this paper, we introduce a deep neural network which represents and generates 3D shapes via {\em sequential part assembly\/}, as shown in 
Figures~\ref{fig:teaser} and~\ref{fig:architecture}. In a way, we regard the assembly sequence as a ``sentence'' which organizes and describes the 
parts constituting a 3D shape. Our approach is inspired, in part, by the resemblance between speech and shape perception, as suggested 
by the seminal work of Biederman~\cite{RBC1987} on recognition-by-components (RBC). Another related observation is that
the phase structure rules for language parsing, first introduced by Noam Chomsky, take on the view that a sentence is both a linear string of words 
and a hierarchical structure with phrases nested in phrases~\cite{borsley1996}. In the context of shape structure presentations, our network 
adheres to linear part orders, while other works~\cite{wang2011,li2017grass,mo2019structurenet} have opted for {\em hierarchical\/} part organizations.  

%

The input to our network is a 3D shape segmented into parts, where each part is first encoded into a feature representation using a part autoencoder;
see Figure~\ref{fig:architecture}(a). The core component of our network is a {\em sequence-to-sequence\/} or {\em Seq2Seq\/} autoencoder which 
encodes a sequence of part features into a latent vector of fixed size, and the decoder reconstructs the 3D shape, one part at a time, resulting in a 
sequential assembly; see Figure~\ref{fig:architecture}(b). With its part-wise Seq2Seq architecture, our network is coined {\em PQ-NET\/}.
The latent space formed by the Seq2Seq encoder enables us to adapt the decoder to perform several generative tasks including shape autoencoding, 
interpolation, new shape generation, and single-view 3D reconstruction, where all the generated shapes are composed of meaningful parts.  

As training data, we take the segmented 3D shapes from PartNet~\cite{Mo_2019_CVPR}, which is built on ShapeNet~\cite{ShapeNet2015}. 
The shape parts are always specified in a file following 
some linear order in the dataset; our network takes the part order that is in a shape file.
We train the part and Seq2Seq autoencoders of PQ-NET separately, either per shape category or across all categories 
of PartNet.

Our part autoencoder adapts IM-NET~\cite{chen2018implicit_decoder} to encode shape parts, rather than whole shapes, with the decoder producing an 
implicit field. The part Seq2Seq autoencoder follows a similar architecture as the original Seq2Seq network developed for machine 
translation~\cite{sutskever2014sequence}. Specifically, the encoder is a bidirectional stacked recurrent neural network (RNN)~\cite{schuster1997bidirectional} that
inputs two sequences of part features, in opposite orders, and outputs a latent vector. The decoder is also a stacked RNN, which decodes the latent vector
representing the whole shape into a sequential part assembly. 

PQ-NET is the first {\em fully generative\/} network which learns a 3D shape representation in the form of sequential part assembly. The only prior part sequence model was 3D-PRNN~\cite{Zou_2017}, which generates part boxes, not their geometry --- our network jointly encodes and decodes part structure and geometry. PQ-NET can be easily adapted to various generative tasks including shape autoencoding, novel shape generation, {\em structured\/} single-view 3D reconstruction from both RGB and depth images, and shape completion. Through extensive experiments, we demonstrate that the performance and output quality of our network is comparable or superior to state-of-the-art generative models including 3D-PRNN~\cite{Zou_2017}, IM-NET~\cite{chen2018implicit_decoder}, and StructureNet~\cite{mo2019structurenet}.



%% file: related.tex

\section{Related work}

\paragraph{Structural analysis of 3D shapes.}
Studies on 3D shape variabilities date back to statistical modeling of human faces~\cite{Blanz:1999:MMS} and bodies~\cite{Allen:2003:SHB},
e.g., using PCA.
Learning \emph{structural variations of man-made shapes} is a more difficult task.
Earlier works from graphics typically infer one or more parametric templates of part arrangement from shape collections~\cite{Ovsjanikov:2011:ECV,Kim:2013:LPT,fish2014meta}.
These methods often require part correspondence of the input shapes.
Probabilistic graphical models can be used to model shape variability
as the causal relations between shape parts~\cite{Kalogerakis:2012:ShapeSynthesis}, but
pre-segmented and part labeled shapes are required for learning such models.




\vspace{-8pt}

\paragraph{``Holistic" generative models of 3D shapes.}
Deep generative models of 3D shapes have been developed for volumetric grids~\cite{wu2016learning,girdhar2016learning,Wang-2017-OCNN,riegler2017octnet}, point clouds~\cite{fan2016point,Achlioptas2018,pointflow}, surface meshes~\cite{groueix2018papier,wang2018pixel2mesh}, multi-view images~\cite{ArsalanCVPR2017},
and implicit functions~\cite{chen2018learning,park2019deepsdf}.
Common to these works is that the shape variability is modeled in a holistic, structure-oblivious fashion.
This is mainly because there are few part-based shape representations suitable for deep learning.

\vspace{-8pt}

\paragraph{Part-based generative models.}
%
%
In recent years, learning deep generative models for part- or structure-aware shape synthesis has been gaining
more interests.
Huang et al.~\cite{Huang:2015:deeplearningsurfaces} propose a deep generative model
based on part-based templates learned \emph{a priori}. 
Nash and Williams~\cite{Nash2017} propose a ShapeVAE to generate segmented 3D objects and
the model is trained using shapes with dense point correspondence.
Li et al.~\cite{li2017grass} propose GRASS, an end-to-end deep generative model of part structures.
They employ recursive neural network (RvNN) to attain hierarchical encoding and decoding of parts and relations.
Their binary-tree-based RvNN is later extended to the N-ary case by StructureNet~\cite{mo2019structurenet}.
Wu et al.~\cite{Wu2018Struct} couple the synthesis of intra-part geometry and inter-part structure.
In G2L~\cite{G2L18}, 3D shapes are generated with part labeling based on generative adversarial networks (GANs) and 
then refined using a pre-trained part refiner. Most recently, Gao et al.~\cite{gao2019sdm} train an autoencoder to 
generate a spatial arrangement of closed, deformable mesh parts respecting the global part structure of a shape category.

Other recent works on part-based generation adopts a generate-and-assemble scheme.
CompoNet~\cite{Schor2019} is a part {\em composition\/} network operating on a fixed number of parts.
Per-part generators and a composition network are trained to produce shapes with a given part structure.
Dubrovina et al.~\cite{Dubrovina2019} propose a decomposer-composer network to learn a factorized shape embedding 
space for part-based modeling. 
Novel shapes are synthesized by randomly sampling and assembling the pre-exiting parts embedded in the factorized latent space.
Li et al.~\cite{li2019pagenet} propose PAGENet which is composed of an array of per-part VAE-GANs, followed by a
part assembly module that estimates a transformation for each part to assemble them into a plausible structure.

\vspace{-8pt}

\paragraph{Seq2Seq.}
Seq2Seq is a general-purpose encoder-decoder framework for machine translation.
It is composed of two RNNs which takes as input a word sequence and maps it into an 
output one with a tag and attention value~\cite{sutskever2014sequence}. To date, 
Seq2Seq has been used for a variety of different applications such as image captioning, 
conversational models, text summarization, as well as few works for 3D representation 
learning. For example, Liu et al.~\cite{liu2019point2sequence} employ Seq2Seq to learn 
features for 3D point clouds with multi-scale context. PQ-NET is the first deep neural
network that exploits the power of sequence-to-sequence translation for generative 
3D shape modeling, by learning structural context within a sequence of constituent shape parts.

\vspace{-8pt}

\paragraph{3D-PRNN: part sequence assembly.}
Most closely related to our work is 3D-PRNN~\cite{Zou_2017}, which, to the best of our knowledge, is the only
prior work that learns a {\em part sequence\/} model for 3D shapes.
Specifically, 3D-PRNN is trained to reconstruct 3D shapes as sequences of {\em box primitives\/} given a 
single depth image. In contrast, our network learns a deep generative model of both a 
linear arrangement of shape parts and geometries of the individual parts. Technically, while both 
networks employ RNNs, PQ-NET learns a shape latent space, jointly encoding both structure and geometry, using 
a Seq2Seq approach. 3D-PRNN, on the other hand, uses the RNN as a recurrent generator that sequentially 
outputs box primitives based on the depth input and the previously generated single primitive. Their network is
trained on segmented shapes whose parts are ordered along the vertical direction.
To allow novel shape generation, 3D-PRNN needs to be initiated by primitive parameters sampled from the training set,
while PQ-NET follows a standard generative procedure using latent GANs~\cite{Achlioptas2018, chen2018implicit_decoder}.

%

\vspace{-8pt}

\paragraph{Single view 3D reconstruction (SVR).}
%
Most methods train convolutional networks that map 2D images to 3D shapes using direct 3D supervision, where voxel~\cite{choy20163d,girdhar2016learning,ogn2017,Matryoshka2018,klokov19bmvc} and point cloud~\cite{fan2017point,mandikal20183d} representations of 3D shapes have been extensively utilized.
Some methods~\cite{lun20173d,arsalan2017synthesizing} learn to produce multi-view depth maps that are fused together into a 3D point cloud.
Tulsiani et al.~\cite{tulsiani2017learning} infer cuboid abstraction of 3D shapes from single-view images. Extending the RvNN-based architecture of GRASS~\cite{li2017grass}, Niu et al.~\cite{niu2018im2struct} propose Im2Struct which maps a single-view image into a hierarchy of part boxes. Differently from this work, our method produces part boxes and the corresponding part geometries jointly, by exploiting the coupling between structure and geometry in a sequential part generative model. 

%% file: method.tex

\section{Method}
\label{sec:method}

In this section, we introduce our \emph{PQ-NET}, based on a \emph{Seq2Seq Autoencoder}, or Seq2SeqAE, for sequential part assembly and part-based shape representation. Given a 3D shape consisting of several parts, we first represents it as a sequence with each vector corresponding to a single part that consists of a geometric feature vector and a 6 DoF bounding box indicating the translating and scaling of part local frame according to the global coordinate system.
The geometry of each part is projected to a low-dimensional feature space based on a hybrid-structure autoencoder using self-supervised training. Since the number of part sequence is un-known, we seek a recurrent neural network based encoder to transform the entire sequence to an unified shape latent space. The part sequence is then decoded from the shape feature vector, with each part containing the geometry feature and the spatial position and size. Figure \ref{fig:architecture} shows the outline of our Seq2SeqAE model. Our learned shape latent space facilitates applications like random generation, single view reconstruction and shape completion, etc. We will explain the two major components of our model in the next sections 
with more details in supplementary material.

\subsection{Part Geometry Auto-encoding} 
The part geometry and topology is much simpler than the original shape. Thus, by decomposing the shape into a set of parts, we are able to perform high-resolution and cross-category geometry learning with high quality. Our part geometry autoencoder uses a similar design as ~\cite{chen2018implicit_decoder}, where a CNN-based encoder projects voxelized part to the part latent space, and a MLP-based decoder re-projects the latent vector to a volumetric Signed Distance Field(SDF). The surface of the object is retrieved using marching cube on the places where SDF is zero.

We first scale each part to a fixed resolution $64\times 64 \times 64$ within its bounding box and feed scaled part volume as input to a CNN encoder to get the output feature vector $g$ that represents the part geometry. The MLP decoder takes in this feature vector $g$ and 3D point $(x, y, z)$ and output a single value that tells either this point is inside the surface of the input geometry or outside. Since volumetric SDF is continuous everywhere, the output geometry is smooth and can be sampled at any resolution. Note that this feature representation has no information about the part's scale and global position, and thus purely captures its geometry property. For a shape with $n$ parts, we can extract a sequence of geometry features $g_1, g_2, ..., g_n$ corresponding to each part.

\subsection{Seq2Seq AE} 
The core of our neural network is a Sequence-to-Sequence(Seq2Seq) Autoencoder. The sequential encoder is a bidirectional stacked RNN~\cite{schuster1997bidirectional} that takes a sequence of part features, along with its reverse version, as the input, and outputs a latent vector $h_z$ of fixed size. This latent vector is then passed to the stacked RNN decoder that outputs a part feature at each time step. Intuitively, the Seq2Seq encoder learns to assemble parts into a complete shape while the decoder learns to decompose it into meaningful parts. In all of our experiments, we used GRU~\cite{cho2014GRU} as the RNN cell and employed two hidden layers for each RNN. 

More specifically, let $F_i = [g_i; b_i]$ denotes part feature vector, concatenated with two components, a part geometry feature $g_i$ and a 6 DoF bounding box $b_i=[x_i, y_i, z_i, l_i, m_i, n_i]$, where $[x_i, y_i, z_i]$ and $[l_i, m_i, n_i]$ indicate box position and size, respectively.
An additional information of part number is used to regularize the shape distribution, since we empirically found it improving the performance. With the extra one-hot vector $t_i$ of part number, the full vector of a part is finally symbolized as $S_i = [F_i;t_i]$.
We feed the sequence of $S=[S_1, S_2, ..., S_n]$ and also its reverse, $S_{reverse}$, to the bidirectional encoder, and obtain two hidden states from the output,

\vspace{-12pt}

\begin{equation}
  \begin{split}
    h_1 = [h_1^1; h_1^2] &= \mathrm{encode_1}(S) \\
    h_2 = [h_2^1; h_2^2] &= \mathrm{encode_2}(S_{reverse}) \\
    h_z &= [h_1^1;h_2^1;h_1^2;h_2^2]
  \end{split}
\end{equation}
The final state $h_z$ is a latent representation of 3D shape.

\begin{figure}[htbp]
\begin{center}
\includegraphics[width=0.99\linewidth, height=0.45\linewidth]{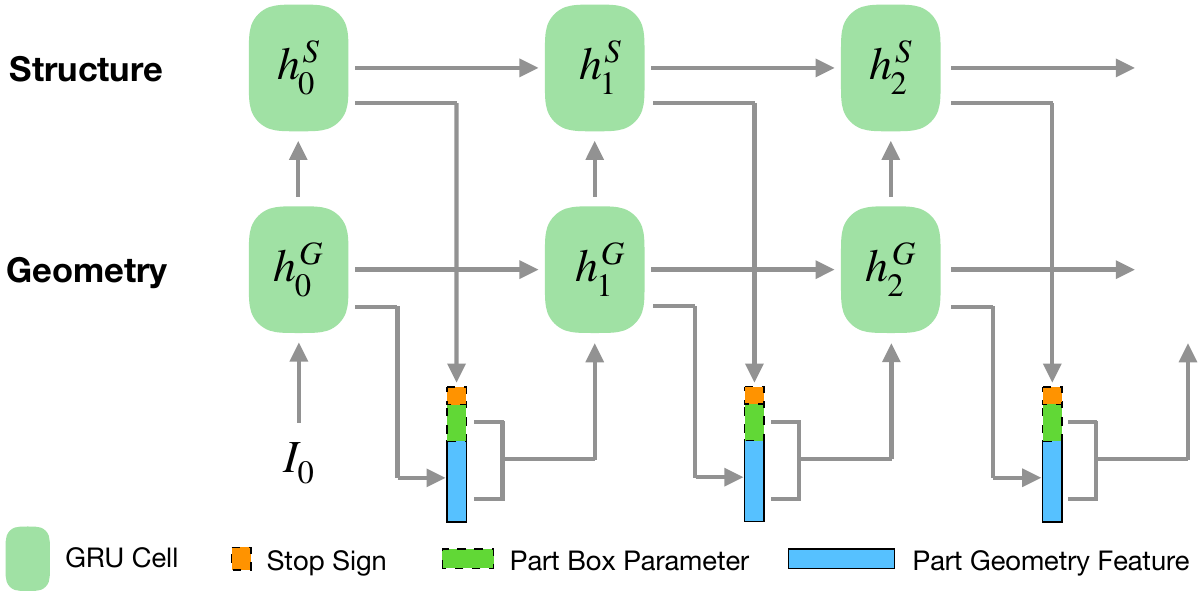}
\end{center}
\vspace{-10pt}
   \caption{Structure of our stacked RNN decoder. At each time step, the geometry feature and structure feature are separately predicted, along with a stop sign indicating whether the iteration is finished.}
\label{fig:stacked-decoder}
\end{figure}

Different to the vanilla RNN, stacked RNN outputs more than one vector for each time step, which allows more complex representation for our parts. Specifically, our stacked RNN has two hidden states at each time step, namely $h_i^G$ and $h_i^S$. We use $h_i^G$ for geometry feature reconstruction by passing it through a MLP network while $h_i^S$ is used for the structure feature with the same technique. We also add another MLP network to predict a stop sign $s_i$ that indicates whether to stop iteration. 
With the initial hidden state set as the final output $h_z$ of encoder RNN, our stacked RNN decoder iteratively generates individual parts by 

\vspace{-10pt}

\begin{equation}
  \begin{split}
    [h_0^S;h_0^G] &= h_z \\
    g_i' &= MLP_G(h_i^G) \\
    b_i' &= MLP_S(h_i^S) \\
    s_i' &= \mathrm{Sigmoid}(MLP_s(h_i^S))
  \end{split}
\end{equation}
The iteration will stop if $s_i' > 0.5$.

Figure \ref{fig:stacked-decoder} illustrates the structure of the RNN decoder. Comparing to the vanilla RNN, where all properties are concatenated into a single feature vector, our disentangling of the geometry and bounding box in a stacked design yields better results without using deeper network.


\subsection{Training and losses}
\label{sec:training_and_losses}
Given a dataset $\mathcal{S}$ with shapes from multiple categories, we describe the training process of our PQ-NET. Due to the complexity of the whole pipeline and the limitation of computational power, we separate the training into two steps. 

\emph{Step 1.} Our part geometry autoencoder consists of a 3D-CNN based encoder $e$ and an implicit function represented decoder $d$. Given a 3D dataset $\mathcal{S}$ with each shape partitioned into several parts, we scale all parts to an unit cube, and collect a 3D parts dataset $\mathcal{P}$. Note that $\mathcal{P}$ is derived from $\mathcal{S}$. 
We use signed distance field for 3D geometry generation as in ~\cite{chen2018implicit_decoder}. Our goal is to train a network to predict the signed distance field of each part $P$ from dataset $\mathcal{P}$. 
Let $T_{P}$ be a set of points sampled from shape $P$, we define the loss function as the mean squared error between ground truth values and predicted values for all points:

\vspace{-5pt}

\begin{equation}
\mathcal{L}(P)= \mathbb{E}_{p \in T_P}{|d(e(P), p) - \mathcal{F}(p)|^2}
\end{equation}
where $\mathcal{F}$ is the ground truth signed distance function.

After the training is done, the encoder $e$ can be used to map each part $P$ to a latent vector $g=e(P)$ which is used as input in the next step. 

\emph{Step 2.} Based on the part sequence representation, we perform jointly analysis of geometry and structure for each shape $S$ using our Seq2Seq model. 
We use a loss function that consists of two parts,

\vspace{-5pt}

\begin{equation}
\label{loss_total}
    \mathcal{L_{\text{total}}} = \mathbb{E}_{S\in\mathcal{S}}[
    \mathcal{L}_{\text{r}}(S) + 
    \alpha \mathcal{L}_\text{stop}(S)
    ],
\end{equation}
where the weighted factor $\alpha$ is empirically set to $0.01$. 

The \emph{reconstruction loss} $\mathcal{L}_r$ punishes the reconstructed geometry and structure feature for being apart to the ground truth. We use mean squared error as the distance measure and define the reconstruction loss as:

\vspace{-5pt}
\begin{equation}
    \mathcal{L}_{\text{r}}(S) = \frac{1}{k}\sum_{i=1}^{k}[\beta ||g'_i - g_i||_2 + ||b'_i - b_i||_2],
\end{equation}
where $k$ is the number of parts of shape $S$, and $\beta$ is set to $1.0$ in our experiments. For the $i$-th part, $g'_i$ and $ b'_i$ denote the reconstructed result of geometry and structural feature while $g_i$ and $b_i$ are the corresponding ground truth.

The \emph{stop loss} $\mathcal{L}_\text{stop}$ encourages the RNN decoder to generate with correct number of parts that exactly fulfills a shape. Similar to 3D-PRNN~\cite{Zou_2017}, we give each time step of RNN decoder a binary label $s_i$ indicating whether to stop at step $i$. 
The stop loss is defined using binary cross entropy:
\begin{equation}
    \mathcal{L}_{\text{stop}}(S) = \frac{1}{k}\sum_{i=1}^{k}[-s_i\log s'_i - (1-s_i)\log (1 - s'_i)]
\end{equation}
where $s'_i$ is the predicted stop sign.

\subsection{Shape Generation and other applications}
\label{subsec:shape_gen}

The latent space learned by PQ-NET supports various applications. We show results of shape auto-encoding, 3D shape generation, interpolation and single-view reconstruction from RGB or depth image in the next section. 

For shape auto-encoding, we use the same setting in the work of ~\cite{chen2018implicit_decoder}.  Each part of a shape is scale to a $64^3$ volume and the point set for SDF regression is sampled around the surface equally from inside and outside. Then the model is trained following the description in Section \ref{sec:training_and_losses}.

For 3D shape generation, we employ latent GANs~\cite{Achlioptas2018, chen2018implicit_decoder} on the pre-learned latent space using our sequential autoencoder. Specifically, we used a simple MLP of three hidden fully-connected layers for both the generator and discriminator, and applied Wasserstein-GAN (WGAN) training strategy with gradient penalty~\cite{pmlr-v70-arjovsky17a, Gulrajani:2017:ITW:3295222.3295327}. After the training is done, the GAN generator maps random vectors sampled from the standard gaussian distribution $\mathcal{N}(0,1)$ to our shape latent space from which our sequential decoder generates new shapes with both geometry and segmentation. 

For 3D reconstruction from single RGB image or depth map, we use a standalone CNN encoder to map the input image to our pre-learned shape latent space. Typically, we use a four convolutional layers CNN as the encoder for depth image embedding and the typical ResNet18 ~\cite{he2016deep} for RGB input embedding. We follow the similar idea as ~\cite{groueix2018papier, chen2018implicit_decoder, mo2019structurenet} to train the CNN encoder while fixing the parameters of our sequential decoder. 


%% file: results.tex

\section{Results, Evaluation, and Applications}
In this section, we show qualitative and quantitative results of our model on several tasks, including shape auto-encoding, shape generation and single view reconstruction. We use PartNet ~\cite{Mo_2019_CVPR}, a large-scale 3D shape dataset with semantic segmentation, in our paper. We mainly use their three largest categories, that is, chair, table and lamp and remove shapes that have more than 10 parts, resulting in 6305 chairs, 7357 tables and 1188 lamps, which are further divided into training, validation and test sets using official data splits of PartNet. 
The original shapes are in mesh representation, and we voxelize them into $64^3$ cube for feature embedding. We follow the sampling approach as in ~\cite{chen2018learning} to collect thousands of 3D point and the corresponding SDF values for implicit field generation. Please refer to our supplementary material for more details on data processing.

\subsection{3D Shape Auto-encoding}
\label{subsec:shape_AE}

\begin{figure}[t!]
\begin{center}
\includegraphics[width=0.99\linewidth, height=0.7\linewidth]{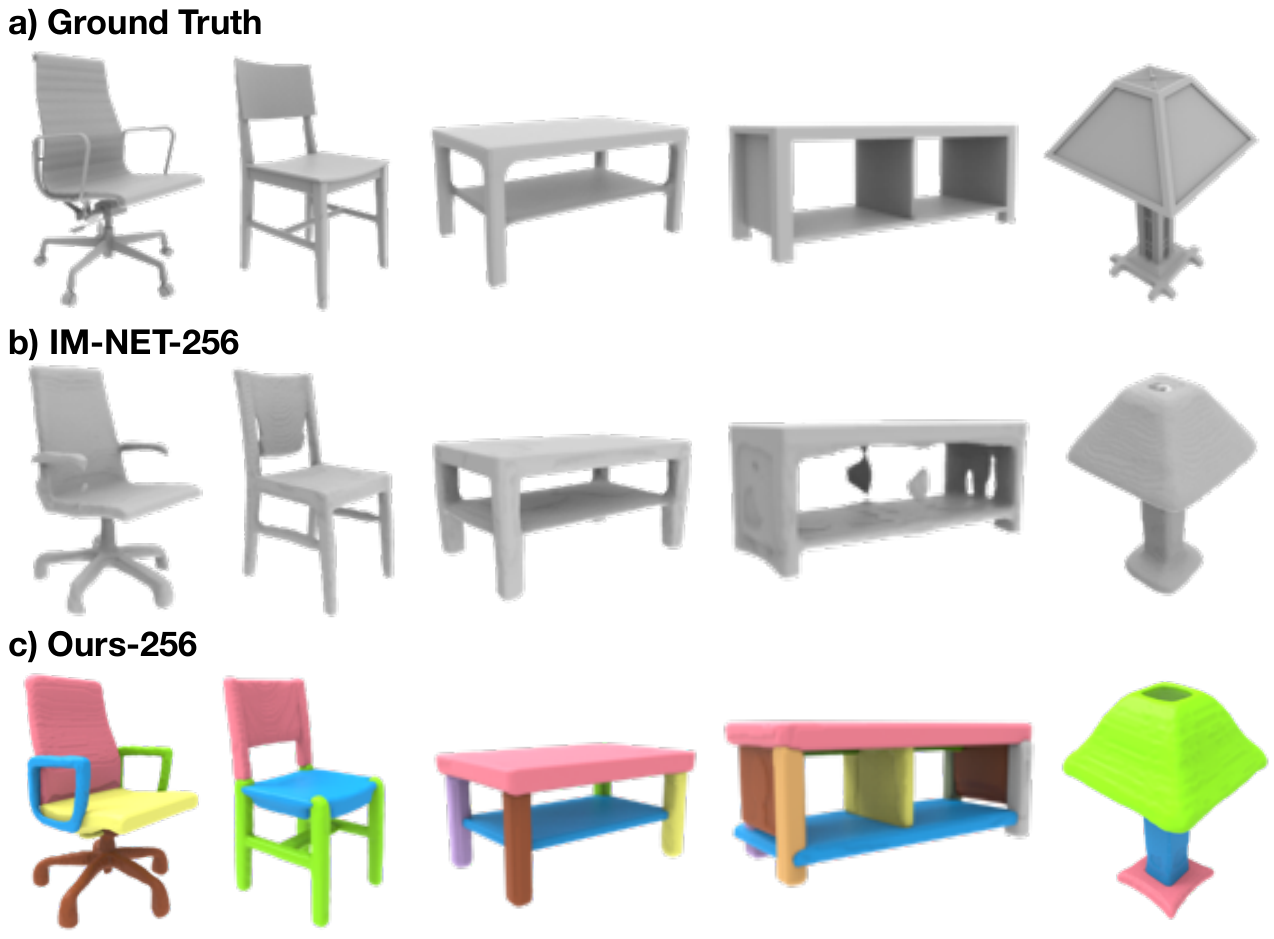}
\end{center}
\vspace{-5pt}
   \caption{Visual results for shape auto-encoding. Output meshes are obtained using the same marching cubes setup.}
\label{fig:reconstruction}
\end{figure}

We compare our sequential autoencoder with IM-NET~\cite{chen2018learning}. Both methods are using the same dataset for training. Table \ref{table:reconstruction} and Figure \ref{fig:reconstruction} shows the results of two methods at different resolutions, specifically $64^3$ and $256^3$. For quantitive evaluation, we use Intersection over Union (IoU), symmetric Chamfer Distance (CD) and Light Field Distance(LFD) ~\cite{LFD} as measurements. 
IoU is calculated at $64^3$ resolution, the same resolution of our training model. 
In Chair category, our method is better than IM-NET, however, in the other two categories, from which the geometry is much simpler, the IoU of IM-NET is better than ours. Note that, the parts of shape generated by our method is better than IM-NET, due to its simplicity, and our generated shape is visually better too. However, small perturbation of part location can significantly cut down the score of IoU.
For CD and LFD, our method performs better than IM-NET. Since LFD is computed within mesh domain, we convert the output of SDF decoder to the mesh using marching cubes algorithm. For CD metric, we samples 10K points on the mesh surface and compare with the ground truth point clouds. 

In general, our model outperforms IM-NET in both qualitative and quantitative evaluation. We admit that this comparison might be a bit unfair for IM-NET, since our inputs are segmented parts, which offers structural information that is not provided by the whole shape. But still the evaluation results show that our model can correctly represents both structure and geometry of 3D shapes. A worth noticing fact is that our cross-category trained model beats per-category trained models. It indicates that our sequential model can handle different arrangements of parts across categories and benefits from the simplicity of part geometry.

\begin{table}[t!]
\begin{center}
\begin{tabular}{|l|l|l|l|l|}
\hline
                 Metrics & Method & Chair & Table & Lamp \\ \hline
\multirow{2}{*}{IoU} 
                  & Ours-64 & \textbf{67.29} & 47.39	 & 39.56 \\ \cline{2-5}
                  & IM-NET-64 & 62.93 & \textbf{56.14} & \textbf{41.29} \\ \hline
\multirow{5}{*}{CD} 
                  & Ours-64 & 3.38 & 5.49 & 11.49 \\ \cline{2-5}
                  & Ours-256 & 2.86 & 5.69 & 10.32 \\ \cline{2-5}
                  & Ours-Cross-256 & \textbf{2.46} & \textbf{4.50} & \textbf{4.87} \\ \cline{2-5}
                  & IM-NET-64 & 3.64 & 6.75 & 12.43 \\ \cline{2-5} 
                  & IM-NET-256 & 3.59 & 6.31 & 12.19 \\ \hline
\multirow{5}{*}{LFD} 
                  & Ours-64 & 2734 & 2824 & 6254 \\ \cline{2-5}
                  & Ours-256 & \textbf{2441} & 2609 & 5941 \\ \cline{2-5}
                  & Ours-Cross-256 & 2501 & \textbf{2415} & \textbf{4875} \\ \cline{2-5}
                  & IM-NET-64 & 2830 & 3446 & 6262 \\ \cline{2-5} 
                  & IM-NET-256 & 2794& 3397 & 6622 \\ \hline
\end{tabular}
\end{center}
\label{table:reconstruction}
\caption{Quantitative shape reconstruction results. IoU is multiplied by $10^2$, CD by $10^3$. LFD is rounded to integer. "Ours-Cross" refers to our model trained across all three categories.}
\end{table}

\subsection{Shape Generation and Interpolation}
\label{subsec:generation}

We compare to two state-of-the-art 3D shape generative models, IM-NET~\cite{chen2018implicit_decoder} and StructureNet~\cite{mo2019structurenet}, for 3D shape generation task. We use the released code for both method. For IM-NET, we retrain their model on all three category.
For StructureNet, we use the pre-trained models on Chair and Table, and retrain the model for Lamp category. 

\begin{table}[t!]
\begin{center}
\begin{tabular}{|l|l|l|l|l|}
\hline
                 Category & Method & COV & MMD & JSD \\ \hline
\multirow{3}{*}{Chair} & Ours & \textbf{54.91} & 8.34 & \textbf{0.0083} \\ \cline{2-5} 
                  & IM-NET & 52.35 & \textbf{7.44} & 0.0084 \\ \cline{2-5}
                  & StructureNet & 29.51 & 9.67 & 0.0477 \\ \hline
\multirow{3}{*}{Table} & Ours & 56.51 & 7.56 & 0.0057 \\ \cline{2-5} 
                  & IM-NET & \textbf{56.67} & \textbf{6.90} & \textbf{0.0047} \\ \cline{2-5}
                  & StructureNet & 16.04 & 14.98 & 0.0725 \\ \hline
\multirow{3}{*}{Lamp} & Ours & \textbf{87.95} & \textbf{10.01} & \textbf{0.0215} \\ \cline{2-5} 
                  & IM-NET & 81.25 & 10.45 & 0.0230 \\ \cline{2-5}
                  & StructureNet & 35.27 & 17.29 & 0.1719 \\ \hline
\end{tabular}
\end{center}
\label{table:quantitative_generation}
\caption{Quantitative evaluation for shape generation. We randomly generated 2000 shapes for each method and then compared to the test dataset. COV and MMD use chamfer distance as distance measure. MMD is multiplied by $10^3$.}
\end{table}

We adopt Coverage (COV), Minimum Matching Distance (MMD) and Jensen-Shannon Divergence (JSD)~\cite{Achlioptas2018} to evaluate the fidelity and diversity of generation results. While COV and JSD roughly represent the diversity of the generated shapes, MMD is often used for fidelity evaluation. We obtained a set of generated shapes for each method by randomly generating 2K samples and compare to the test set using chamfer distance. 
More details about evaluation metrics are available in supplementary material.



The results of PQ-NET and IM-NET are sampled at resolution $256^3$ for visual comparison and $64^3$ for quantitative evaluation. We reconstruct the mesh and sample 2K points to calculate chamfer distance. Since StructureNet outputs 1K points for each generated part, the whole shape may contain points larger than 2K. We conduct a downsampling process to extract 2K points for evaluation.

\begin{figure}[t!]
\begin{center}
\includegraphics[width=0.99\linewidth, height=0.7\linewidth]{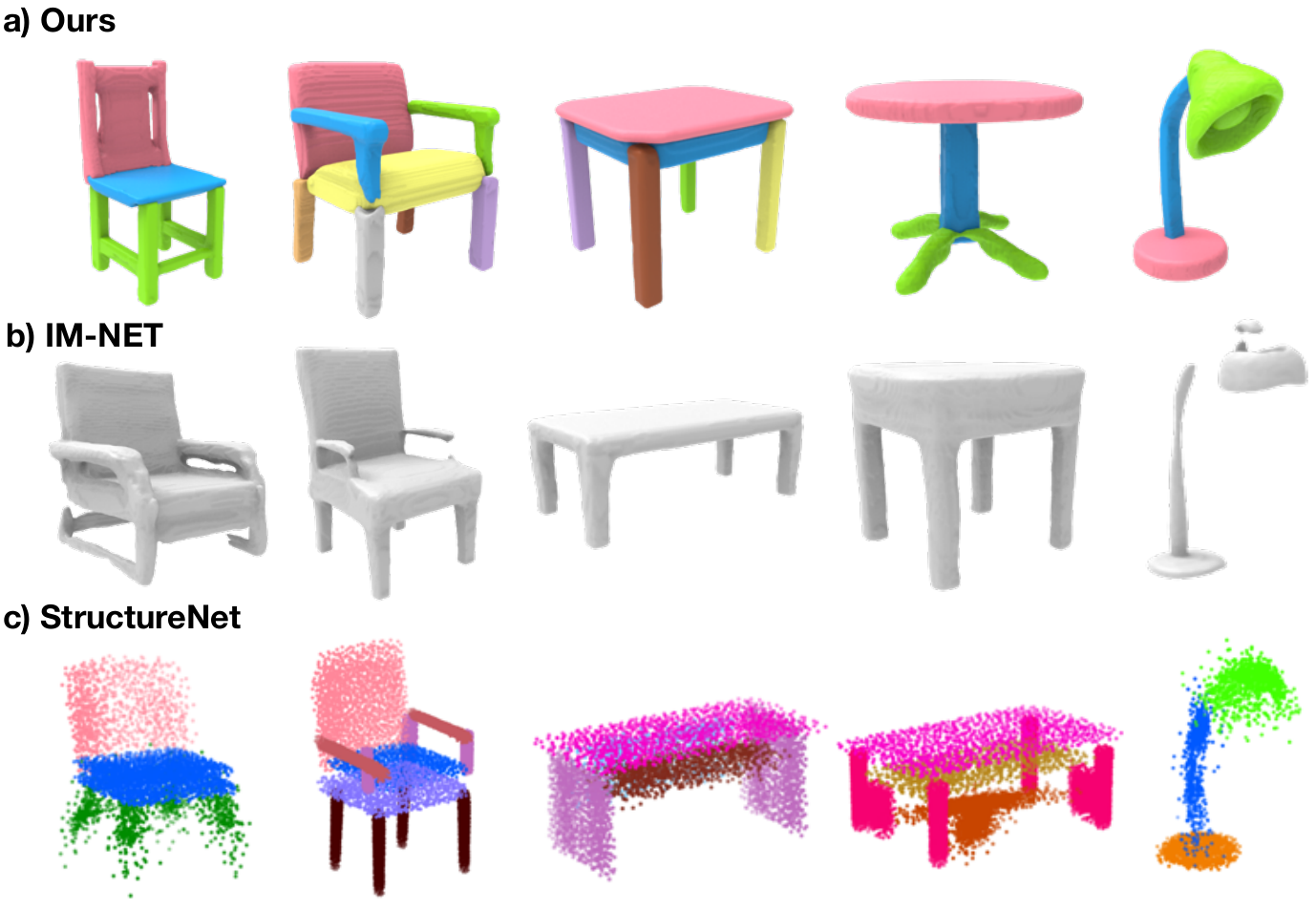}
\end{center}
\vspace{-5pt}
   \caption{3D shape generation results with comparison to results obtained by IM-NET and StructureNET.}
\label{fig:generation-compare}
\end{figure}

\begin{figure}[t!]
\begin{center}
\includegraphics[width=0.99\linewidth]{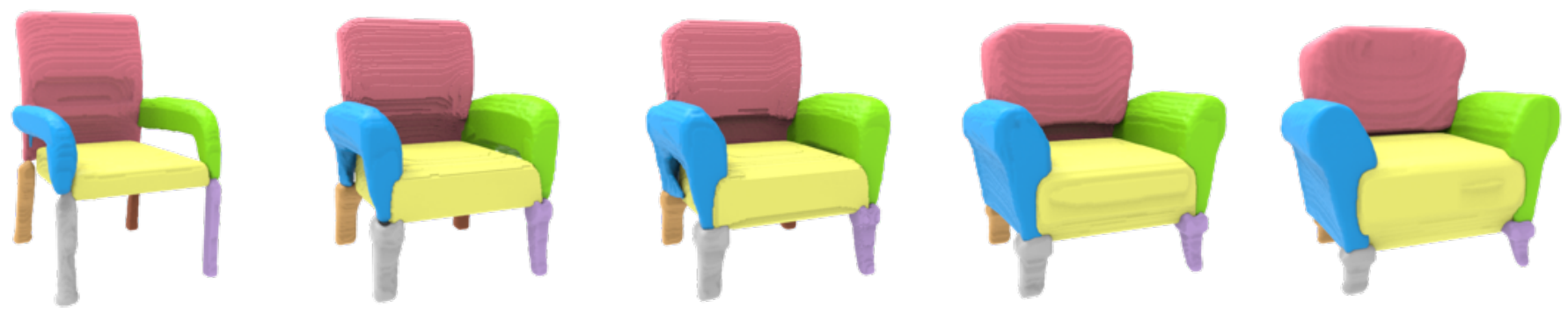}
\end{center}
\vspace{-5pt}
   \caption{Latent space interpolation results. The interpolated sequence not only consists of smooth geometry morphing but also keeps the shape structure.}
\label{fig:interpolation}
\end{figure}

Table \ref{table:quantitative_generation} and Figure \ref{fig:generation-compare} shows the results from our PQ-NET, IM-NET and StructureNet. Our method can produce smooth geometry while maintaining the whole structure preserved. For thin structure and complex topology, modeling whole shape is very hard, and our decomposition strategy can be very helpful in such hard situation. However, on the other hand, our sequential model may yield duplicated parts or miss parts sometimes. As to get the sufficient generative model, it is important to balance the hardness between geometry generation and structure recovery.

Besides random generation, we also show interpolation results in Figure \ref{fig:interpolation}. Interpolation between latent vector is a way to show the continuity of learned shape latent space. Linear interpolation from our latent space yields smooth transiting shapes in terms of geometry and structure.

\subsection{Comparison to 3D-PRNN}
\label{subsec:comp_3DPRNN}

Since 3D-PRNN~\cite{Zou_2017} is the most related work, we conduct a comprehensive comparison with them.
We first compare the reconstruction task from a single depth image by evaluating only the structure of shape, since 3D-PRNN doesn't recover shape geometry. For each 3D shape in the dataset, we obtain 5 depth maps by the resolution of $64^2$. We uniformly sample 5 views and render the depth images using ground truth mesh. For both 3D-PRNN and our model, we use part axis aligned bounding box(AABB) as structure representation. In addition, 3D-PRNN uses a pre-sort order from the input parts. Therefore, besides using the natural order from PartNet annotations, we also train the model on the top-town order used by 3D-PRNN. 

\begin{figure}[t!]
\begin{center}
\includegraphics[width=0.99\linewidth, height=0.55\linewidth]{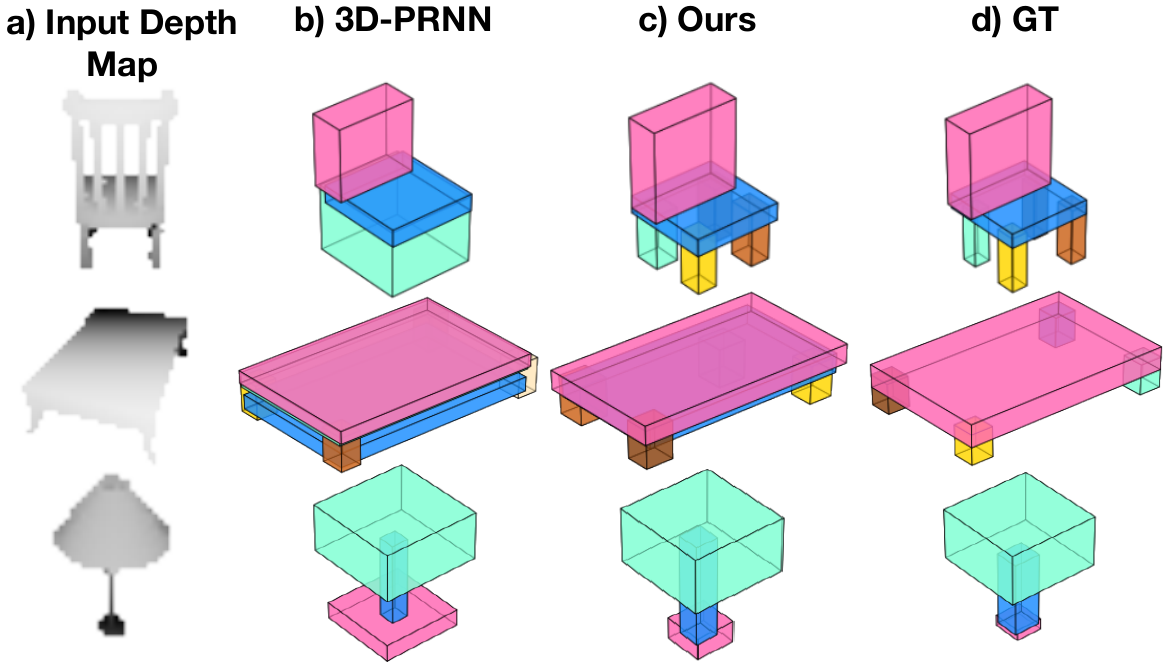}
\end{center}
\vspace{-5pt}
   \caption{Visual comparison of structured 3D shape reconstruction from single depth image on three categories: chair, table, lamp. }
\label{fig:3dprnn-depthsvr}
\end{figure}

Figure \ref{fig:3dprnn-depthsvr} shows the visual comparison between our PQ-NET and 3D-PRNN. Our method can reconstruct much plausible boxes. For quantitative evaluation, we convert the output and ground truth boxes to volumetric model by fully filling with each part box, and compute IoU between generated model and the corresponding ground truth volume. As a result, our reconstructed structures are more accurate, as shown in Table \ref{table:3dprnn-depthsvr}. 
In terms of order effect, our model on the natural order of PartNet yields the best result. The quality drops down with small portion when using the top-down order as 3D-PRNN, however is still better than theirs. 

\begin{table}[t!]
\begin{center}
\begin{tabular}[width=0.99\linewidth]{|l|l|l|l|l|l|}
\hline
Method & Order & Chair & Table & Lamp & Average\\ \hline
\multirow{2}{*}{Ours} & A &\textbf{61.47} & \textbf{53.67} & \textbf{52.94} & \textbf{56.03} \\ \cline{2-6} 
					  & B & 58.68 & 48.58 & 52.17 & 53.14 \\ \hline
\multirow{2}{*}{3D-PRNN} & A & 37.26 & 51.30 & 47.26 &  45.27 \\ \cline{2-6} 
						 & B & 36.46 & 51.93 & 43.83 & 44.07 \\ \hline 

\end{tabular}
\end{center}
\label{table:3dprnn-depthsvr}
\caption{Shape IoU evaluation of structured 3D shape reconstruction from single depth image on three categories: chair, table, lamp. We test each method on two kinds of order: PartNet natural order(A) and presorted top-down order(B).}
\end{table}

We also compare the 3D shape generation task with 3D-PRNN, as shown in Figure \ref{fig:3dprnn-generation}. Quantitative evaluation and more details can be found in supplementary material.

\vspace{-5pt}

\begin{figure}[t!]
\begin{center}
\includegraphics[width=1.0\linewidth, height=0.4\linewidth]{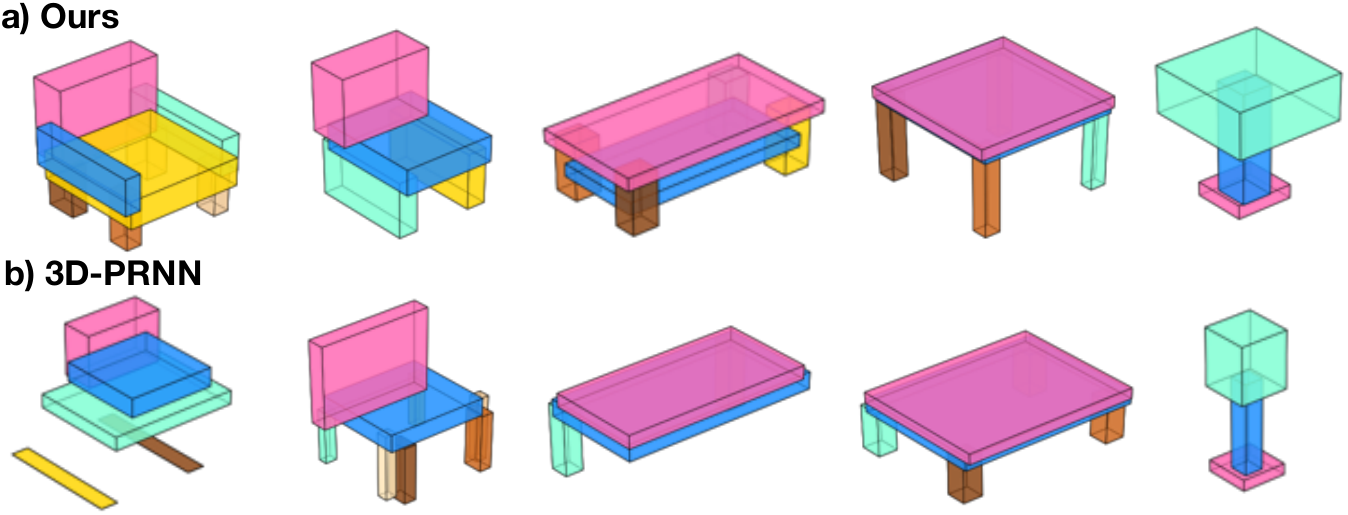}
\end{center}
\vspace{-10pt}
   \caption{Visual comparison of random generated 3D primitives. 3D-PRNN suffers from unreal, duplicated or missing parts while our model can yield more plausible results.}
\label{fig:3dprnn-generation}
\end{figure}

%

\subsection{Single View 3D Reconstruction}	
\label{subsec:SVR}

\begin{figure}[t!]
\begin{center}
\includegraphics[width=\linewidth, height=0.8\linewidth]{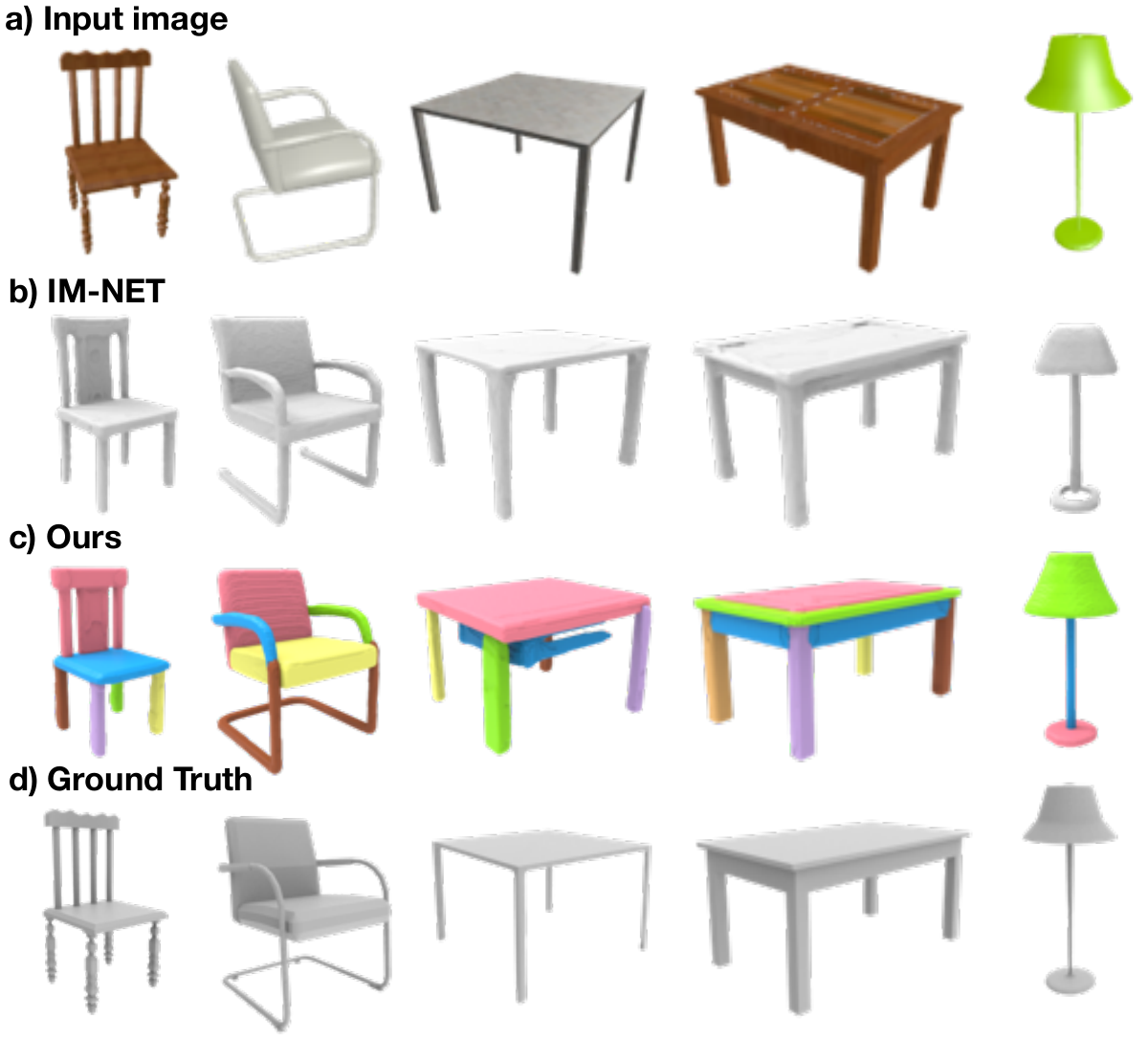}
\end{center}
\vspace{-10pt}
   \caption{Single view reconstruction results. Our results are from model that is trained across all three category. Note that our method also recovers the shape structure. }
\vspace{-5pt}
\label{fig:svr-compare}
\end{figure}

We compare our approach with IM-NET ~\cite{chen2018learning} on the task of single view reconstruction from RGB image. We per-category trained IM-NET on PartNet dataset. Figure \ref{fig:svr-compare} shows the results. It can be seen that our approach can recover more complete and detailed geometry than IM-NET. The advantage of model is that we also obtain segmentation besides reconstructed geometry. However, relying on the structure information may cause issues, such as duplicated or misplaced part, see the first table in Figure \ref{fig:svr-compare}(c). 

We admit our method doesn't outperform IM-NET in the quantitative evaluation. This may due to the fact that our latent space is entangled with both the geometry and structure, which makes the latent space less uniform.

%% file: application.tex
\subsection{Applications}
\label{subsec:apps}

By altering the training procedure applied to our network, we show that PQ-NET can serve two
more applications which benefit from sequential part assembly.

\vspace{-12	pt}

\paragraph{Shape completion.}
We can train our network by feeding it input part sequences which constitute a {\em partial\/} shape, and force the network
to reconstruct the full sequence, hence completing the shape. We tested this idea on the chair category, by randomly removing up to $k - 1$ parts from the part sequence, $k$ being the total number of parts of a given shape. One result is shown in Figure~\ref{fig:teaser} with more available in the supplementary material.


\vspace{-12pt}

\begin{figure}[t!]
\begin{center}
\includegraphics[width=1.0\linewidth, height=0.6\linewidth]{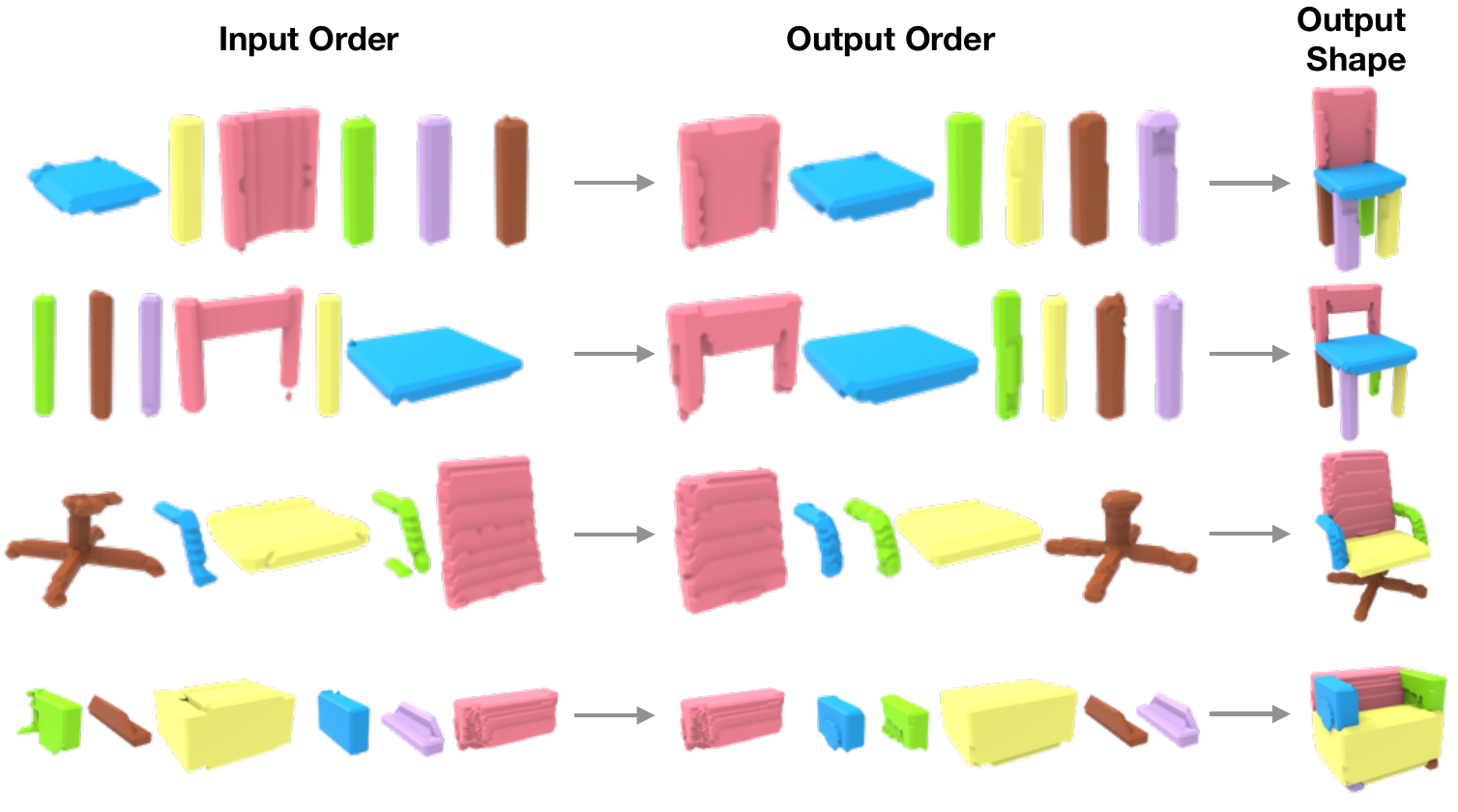}
\end{center}
\vspace{-10pt}
   \caption{Part order denoising results. Our method can unscramble random input orders into a consistent output order, to facilitate part correspondence. Note that the color correspondence is for illustrations only, and not part of the output from our network.}
\label{fig:order-denoising}
\end{figure}

\paragraph{Order denoising and part correspondence.}
We can add ``noise'' to a part order by scrambling it, feed the resulting noisy order to our network, and force it
to reconstruct the original (clean) order. We call this procedure {\em part order denoising\/} --- it
allows the network to learn a {\em consistent\/} part order for a given object category, e.g., chairs, as long as we
provide the ground truth orders with consistency. For example, we can enforce the order 
``back $\rightarrow$ seat $\rightarrow$ legs'' and for the legs, we order them in clockwise order. If all the part
orders adhere to this, then it should be straightforward to imply a {\em part correspondence\/}, which can, in turn,
facilitate inference of part relations such as symmetry; see Figure~\ref{fig:order-denoising}.

With structural variaties, it still requires some work to infer the part
correspondence from all possible (consistent) linear part sequences; this is beyond the scope of our current work.
It is worth noting however that this inference problem would be a lot harder if the parts are organized hierarchically~\cite{wang2011,li2017grass,mo2019structurenet} rather than linearly. 

%% file: conclusion.tex
\section{Conclusion, limitation, and future work}

We present PQ-NET, a deep neural network which represents and generates 3D shapes as an assembly sequence of parts. 
The generation can be from random noise to obtain novel shapes or conditioned on single-view depth scans or RGB 
images for 3D reconstruction. Promising results are demonstrated for various applications and in comparison with 
state-of-the-art generative models of 3D shapes including IM-NET~\cite{chen2018implicit_decoder}, StructureNet~\cite{mo2019structurenet}, and 3D-PRNN~\cite{Zou_2017}, where the latter work also generates part assemblies.

One key limitation of PQ-NET is that it does not learn part {\em relations\/} such as symmetry; it only outputs a spatial arrangement of shape parts. More expressive structural representations such as symmetry hierarchies~\cite{wang2011,li2017grass} and graphs~\cite{mo2019structurenet} can encode such relations easily. However, to learn such representations, one needs to prepare sufficient training data which is a non-trivial task. The part correspondence application shown in Section~\ref{subsec:apps} highlights an advantage of sequential representations, but in general, an investigation into the pros and cons of sequences vs.~hierarchies for learning generative shape models is worthwhile.
Another limitation is that PQ-NET does not produce topology-altering interpolation, especially between shapes with different number of parts. Further investigation into latent space formed by sequential model is needed. 

We would also like to study more closely the latent space learned by our network, which seems to be encoding part structure
and geometry in an entangled and unpredictable manner. This might explain in part why the 3D reconstruction quality from PQ-NET still does not quite match that of state-of-the-art implicit models such as IM-NET. Finally, as shown in Table~\ref{table:3dprnn-depthsvr}, part orders do seem to impact the network learning. Hence, rather than adhering to a fixed part order, the network may learn a good, if not the optimal, part order, for different shape categories, i.e., the best assembly sequence. An intriguing question is what would be an appropriate loss to quantify the best part order.

%% file: supp.tex
\textbf{\large Supplementary Materials}

\setcounter{section}{0}
 \renewcommand{\thesection}{\Alph{section}}


\section{Overview}
This supplementary material contains six parts:
\vspace{-7pt}
\begin{itemize}
	\item Sec.\ref{sec:implementation} describes the implementation detailed of our PQ-NET.
\vspace{-7pt}
	\item Sec.\ref{sec:data} describes the data preparation details.
\vspace{-7pt}
	\item Sec.\ref{sec:metrics} explains the metrics used for the evaluation of shape generation.
\vspace{-7pt}
	\item Sec.\ref{sec:3dprnn} provides comparison results to 3D-PRNN on generation task.
\vspace{-7pt}
	\item Sec.\ref{sec:more-results} provides more visual results of partial shape completion, random shape generation.
\end{itemize}

\input{supps-section/implementation}

\input{supps-section/dataprocess}

\input{supps-section/metrics}

\input{supps-section/3dprnn}

\section{More Results}
\label{sec:more-results}
\begin{figure}[hb]
\begin{center}
\includegraphics[width=\linewidth, height=0.8\linewidth]{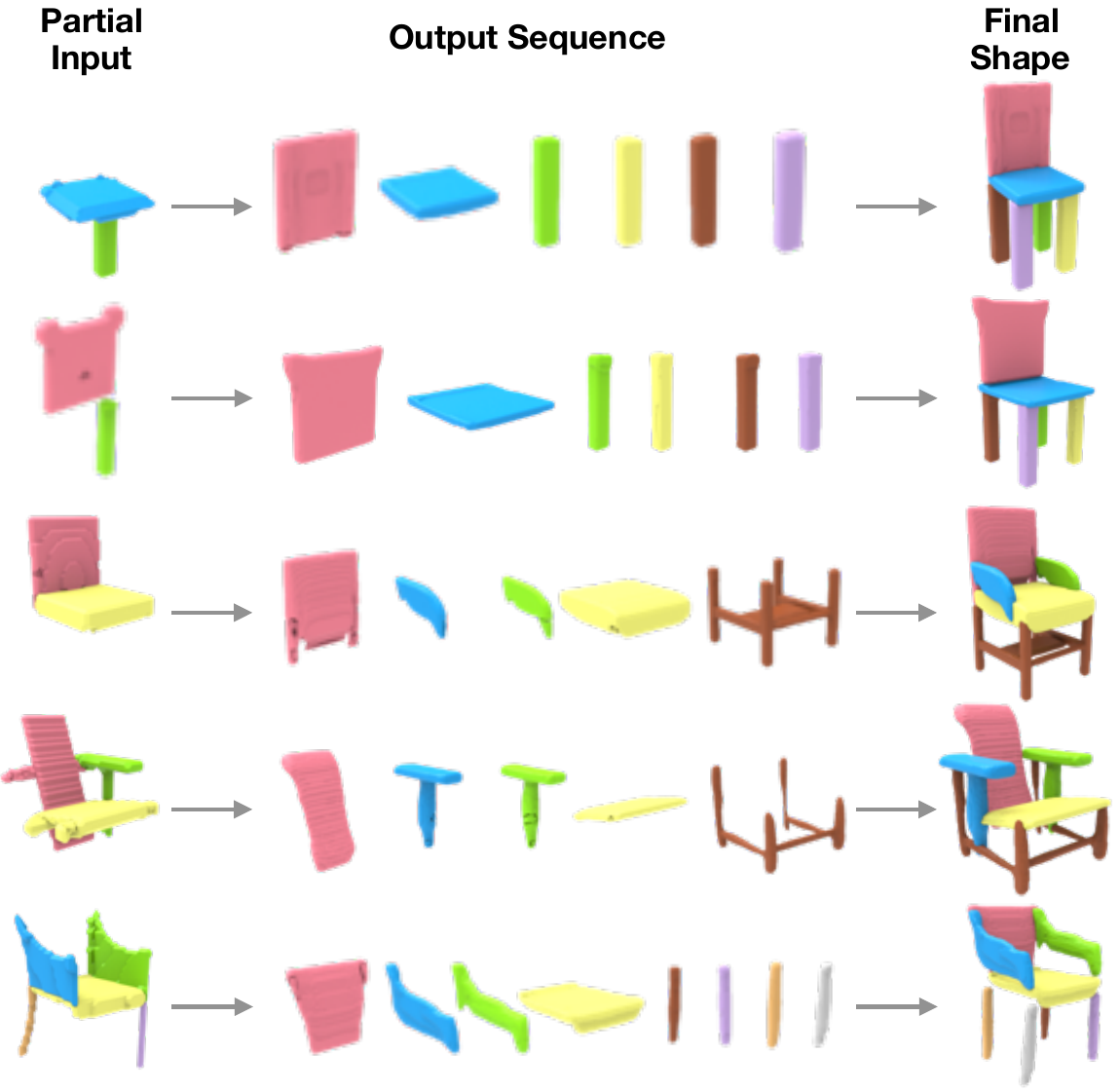}
\end{center}
\vspace{-5pt}
   \caption{Visual results of partial shape completion.}
\label{fig:complement-more}
\end{figure}

\begin{figure*}[htbp]
\begin{center}
\includegraphics[width=\textwidth, height=1.2\textwidth]{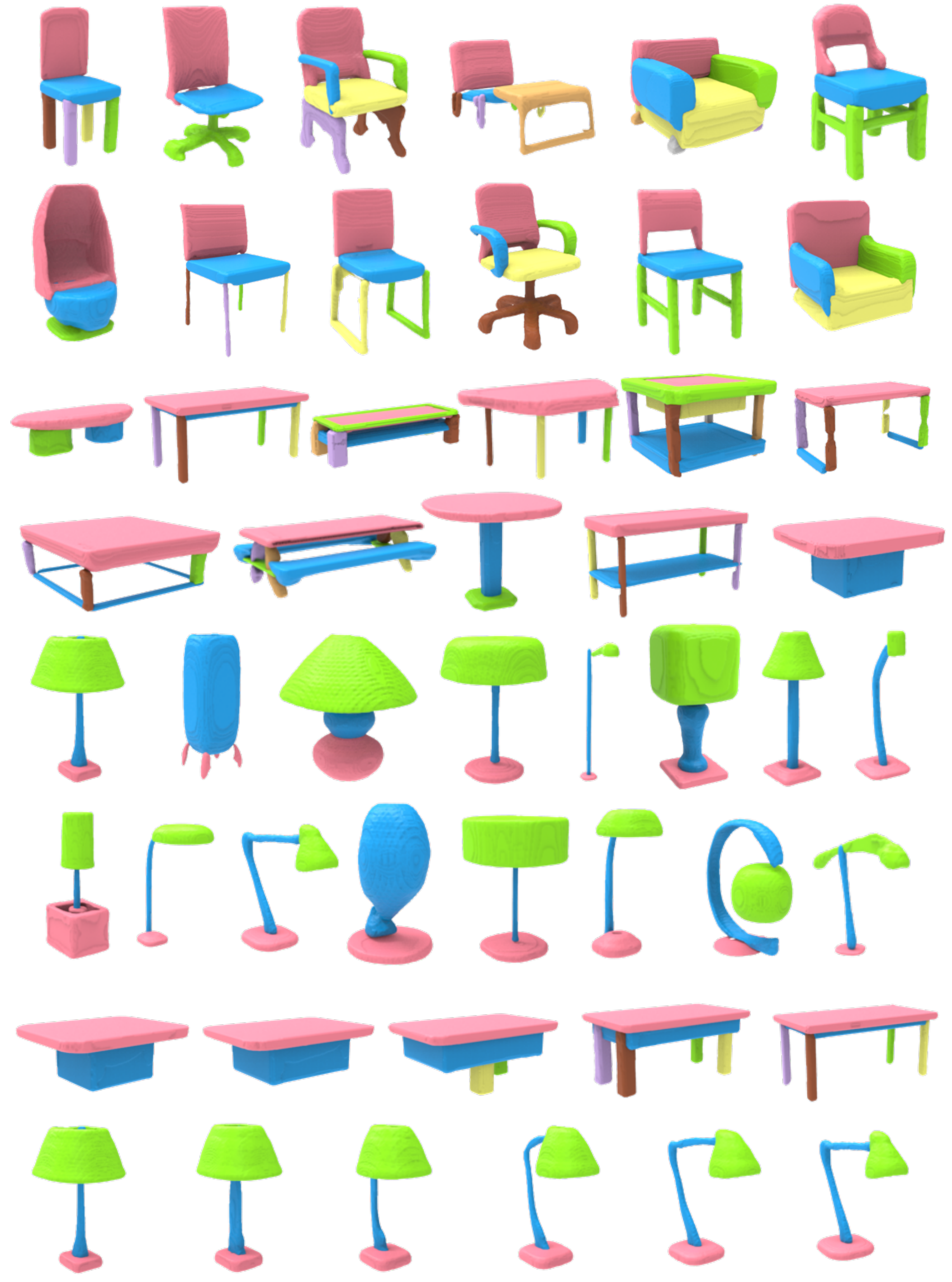}
\end{center}
\caption{More visual results of our generated shapes (row 1-6), along with two latent space interpolation (row 7-8). All shapes are sampled at resolution $256^3$ and reconstructed using Marching Cubes.}
\label{fig:generation-more}
\end{figure*}

Figure \ref{fig:complement-more} shows visual results for partial shape completion and Figure \ref{fig:generation-more} at the last page shows more results of our generated shapes.

\newpage

%% file: supps-section/implementation.tex

\section{Implementation Details}
\label{sec:implementation}
We describe the detailed implementation of our PQ-NET architecture along with the training configuration. 
\begin{table}[h]
\begin{center}
\resizebox{\linewidth}{!}{
\begin{tabular}{|l|l|l|l|}

\hline
\multicolumn{4}{|c|}{\bf CNN Encoder} \\ \hline
Layer & Kernel Size & Stride  & Output Shape \\ \hline
input voxel & - & - & (1,64,64,64) \\ \hline
Conv3D+BN+lReLU & (4,4,4) & (2,2,2) & (32,32,32,32) \\ \hline
Conv3D+BN+lReLU & (4,4,4) & (2,2,2) & (64,16,16,16) \\ \hline
Conv3D+BN+lReLU & (4,4,4) & (2,2,2) & (128,8,8,8) \\ \hline
Conv3D+BN+lReLU & (4,4,4) & (2,2,2) & (256,4,4,4) \\ \hline
Conv3D+Sigmoid & (4,4,4) & (1,1,1) & (128,1,1,1) \\ \hline
\multicolumn{4}{|c|}{\bf Implicit Decoder} \\ \hline
Layer & Dropout & Input Shape  & Output Shape \\ \hline
feature+coordinates & - & (128 + 3) & (131) \\ \hline
FC+lReLU & 0.4 & (131) & (2048) \\ \hline
FC+lReLU & 0.4 & (2048 + 131) & (1024) \\ \hline
FC+lReLU & 0.4 & (1024 + 131) & (512) \\ \hline
FC+lReLU & 0.4 & (512 + 131) & (256) \\ \hline
FC+lReLU & - & (256 + 131) & (128) \\ \hline
FC+Sigmoid & - & (128) & (1) \\ \hline

\end{tabular}}
\vspace{+1pt}
\caption{Architecture of our part geometry autoencoder. Conv3D: 3D Convolutional Layer, BN: Batch Normalization, lReLU: leaky ReLU, FC: Fully Connected Layer.}
\label{table:detail_imnet}

\vspace{+5pt}

\resizebox{\linewidth}{!}{\begin{tabular}{|l|l|l|l|l|}
\hline
\multicolumn{5}{|c|}{\bf Seq2Seq Encoder RNN} \\ \hline
Type & \#layers & input size  & hidden size & bidirectional \\ \hline
GRU & 2 & (128+6+K) & 256 & True \\ \hline
\multicolumn{5}{|c|}{\bf Seq2Seq Decoder RNN} \\ \hline
Type & \#layers & input size  & hidden size & bidirectional \\ \hline
GRU & 2 & (128+6) & 512 & False \\ \hline
\multicolumn{5}{|c|}{\bf Seq2Seq Decoder FC} \\ \hline
\multicolumn{2}{|c|}{Type} & input size  & hidden size & output size \\ \hline
\multicolumn{2}{|c|}{FC-lReLU-FC}  & 512 & 256 & 128\\ \hline
\multicolumn{2}{|c|}{FC-ReLU-FC}  & 512 & 128 & 6\\ \hline
\multicolumn{2}{|c|}{FC-ReLU-FC}  & 512 & 128 & 1\\ \hline

\end{tabular}
}
\vspace{+1pt}
\caption{Architecture of our Seq2Seq autoencoder. K is the maximum number of parts of all shaeps in the dataset.}
\label{table:detail_seq2seq}
\end{center}
\end{table}

Table \ref{table:detail_imnet} and Table \ref{table:detail_seq2seq} list the detailed architecture with specific parameters of our PQ-NET, divided into part geometry autoencoder and Seq2Seq autoencoder. For part geometry autoencoder, we use similar design as IM-NET ~\cite{chen2018implicit_decoder}, with skip connection in the implicit decoder. For Seq2Seq autoencoder, we also employ dropout regularization with drop rate $0.2$ in the middle of GRU to reduce overfitting.

As mentioned in the main paper, we train our PQ-NET in two separate steps. We adopt a progressive strategy to train our part geometry autoencoder by increasing the resolution of part volum. Practically, we use resolution of $16^3$, $32^3$ and $64^3$ with batch size $40$ and learning rate 5e-4 in our experiments. Then the part geometry autoencoder is fixed and used to train the Seq2Seq autoencoder on the resolution of $64^3$ with batch size as 64 and learning rate as 1e-3. We use PyTorch~\cite{paszke2017automatic} framework to implement our PQ-NET and conduct all the experiments.

%% file: supps-section/dataprocess.tex

\section{Data Preparation Details}
\label{sec:data}
For all experiments in our paper, we mainly use three largest categories of PartNet~\cite{Mo_2019_CVPR}, that is, chair, table and lamp. Since each shape in PartNet is partitioned into small elements and then grouped following a hierarchical structure with each node a semantical label, we use the nodes in the second layer as our part geometry. The part label we used appears in the file "partnet-dataset$\backslash$stats$\backslash$after\_merging\_label\_ids$\backslash$xxx-label-2.txt", where "xxx" correspondes to the number of each shape categories. We remove shapes that contain more than 10 parts, resulting in 6305 chairs, 7357 tables and 1188 lamps, which are further divided into training, validation and test sets using official data splits of PartNet. Note that this upper bound of number of parts can be increased. Table \ref{table:part_dataset} shows the statistics result about the number of parts per shape in our dataset.

\begin{table}[htbp]
\begin{center}
\begin{tabular}{|l|l|l|l|}
\hline
 & Chair & Table & Lamp \\ \hline
avg \#parts & 5.59 & 6.06 & 2.94 \\ \hline
min \#parts & 2 & 2 & 2 \\ \hline
max \#parts & 9 & 10 & 7 \\ \hline
\end{tabular}
\end{center}
\vspace{-5pt}
\caption{Average, minimum and maximum number of parts per shape for each category in our dataset.}
\label{table:part_dataset}
\end{table}


To prepare the training data for our network, we first voxelize the original shape mesh into voxel representation at $64^3$ resolution and fill the interior of the shape voxel using classic flood filling algorithm. Each part is then scaled to $64^3$ resolution within its bounding box. The resulting $64^3$ part volums are downsampled to $32^3$ and $16^3$ resolution. With voxelized part geometry at differ resolutions, we follow the sampling approach as in IM-NET to progressively sample points near the surface with each point a signed distance to the surface. Specifically, we sample $4096$ points in $16^3$ volum, and with higher resolutions, such as $32^3$ and $64^3$, we sample $8192$ and $32768$, respectively. The sampled points together with signed distance values are used to train our part geometry autoencoder. The box parameters used in Seq2Seq autoencoder training are produced by calculating bounding box of each part in original shape voxel. Intuitively, the box parameters indicate the deviation and translation from part local frame to the shape coordinate system.

%% file: supps-section/metrics.tex

\section{Metrics}
\label{sec:metrics}
We explain the quantitative metrics adopt for generation task in our paper, \ie Coverage (COV), Minimum Matching Distance (MMD) and Jensen-Shannon Divergence (JSD)~\cite{}. In their calculation, chamfer distance is used when comparing our method to IM-NET~\cite{chen2018implicit_decoder} and StructureNet~\cite{mo2019structurenet} while IoU is used when comparing to 3D-PRNN~\cite{Zou_2017}. Let $\mathcal{G}$ be a set of generated shapes and $\mathcal{S}$ be the ground truth test set:
\vspace{-13pt}
\paragraph{COV} To compute COV, for each shape in $\mathcal{G}$ we find its nearest neighbor in $\mathcal{S}$ and mark it as matched. COV is the fraction of matched shapes in $\mathcal{S}$ over the total size of $\mathcal{S}$. COV roughly represents the diversity of the generated shapes. A high COV score suggests most of shapes in $\mathcal{S}$ can be roughly represented by shapes in $\mathcal{G}$.
\vspace{-14pt}
\paragraph{MMD} To compute MMD, for each shape in $\mathcal{S}$, we calculate the distance to its nearest neighbor in $\mathcal{G}$. Then MMD is defined as the average of all these distances. MMD roughly represents the fidelity of the generated shapes.
\vspace{-14pt}
\paragraph{JSD} In a predefined voxel gird, for each shape in point cloud form in $\mathcal{G}$, we count the number of points lying inside each voxel, and do the same for $\mathcal{S}$. Then we get two distribution in Euclidean 3D space $P_{\mathcal{G}}$ and $P_{\mathcal{S}}$. JSD is defined as the Jensen-Shannon Divergence between two distributions.

We use the code from \url{https://github.com/optas/latent_3d_points} for calculation.

%% file: supps-section/3dprnn.tex

\section{Comparison to 3D-PRNN on Shape Generation Task}
\label{sec:3dprnn}
In this section, we demonstrate detailed comparison results to 3D-PRNN~\cite{Zou_2017} on shape generation task, which are not fully shown in the main paper due to paper length limitation. 
Unlike our PQ-NET that generates new shapes from random noise, 3D-PRNN samples new structure within a constraint region. For a fair comparison, we follow the setting described in their paper, by sampling the first input feature of RNN from training data. 

\begin{table}[t]
\begin{center}
\resizebox{\linewidth}{!}{
\begin{tabular}{|l|l|l|l|l|l|}
\hline
Metrics & Method & Chair & Table & Lamp & Avg \\ \hline
\multirow{2}{*}{COV-IoU} & Ours & \textbf{58.94} & \textbf{59.80} & \textbf{76.34} & \textbf{65.03} \\ \cline{2-6} 
						& 3D-PRNN & 51.03 & 36.44 & 58.93 & 48.80\\  \hline
\multirow{2}{*}{MMD-IoU} & Ours & \textbf{0.259} & \textbf{0.271} & \textbf{0.298} & \textbf{0.276} \\ \cline{2-6} 
						& 3D-PRNN & 0.275 & 0.357 & 0.347 & 0.326\\ \hline
\end{tabular}
}
\end{center}
\vspace{-5pt}
\caption{Quantitative comparison on 3D shape generation.}
\label{table:3dprnn-generation}
\end{table}

\begin{figure}[t]
\begin{center}
\includegraphics[width=0.9\linewidth, height=0.8\linewidth]{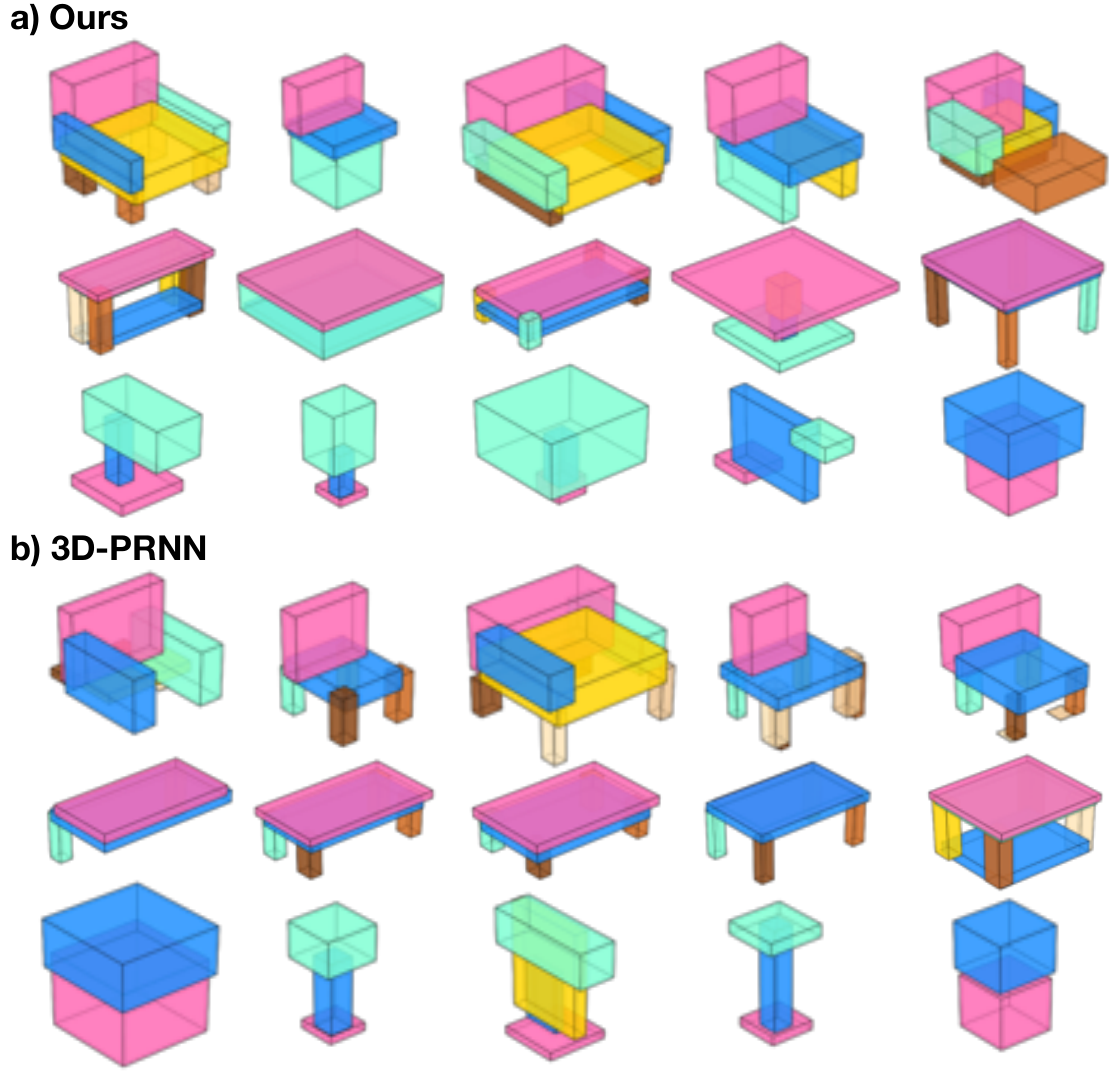}
\end{center}
\vspace{-10pt}
   \caption{More visual comparison of random generated 3D primitives between ours and 3D-PRNN.}
\label{fig:3dprnn-generation-more}
\end{figure}

Results of quantitative comparison on are shown in and Table \ref{table:3dprnn-generation}. We sampled 2000 random generated shapes for chair and table, 800 for lamp, to compute the coverage(COV) and minimum matching distance(MMD) between the generated set and ground truth test set. We use $1 - \text{IoU}$ as the distance measure when comparing two shapes. It can be seen that our network outperforms 3D-PRNN in all of the measurements, which means that our generation results are more diverse and plausible. More visual results are shown in Figure \ref{fig:3dprnn-generation-more}.